\documentclass{article}
\usepackage{spconf,amsmath,graphicx}
\usepackage{textcomp} 
\usepackage{amssymb} 
\usepackage{textcomp} 
\usepackage{amsmath}    
\usepackage{tabularx}   
\usepackage[skip=3pt]{caption}  

\title{Vision-language cross-attention for real-time autonomous driving}

\name{Santosh Patapati, Trisanth Srinivasan, Murari Ambati}
\address{Cyrion Labs}

\begin{document}
%
\maketitle
\begin{abstract}
Autonomous cars need geometric accuracy and semantic understanding to navigate complex environments, yet most stacks handle them separately. We present XYZ-Drive, a single vision-language model that reads a front-camera frame, a 25m × 25m overhead map, and the next waypoint, then outputs steering and speed. A lightweight goal-centered cross-attention layer lets waypoint tokens highlight relevant image and map patches, supporting both action and textual explanations, before the fused tokens enter a partially fine-tuned LLaMA-3.2 11B model.

On the MD-NEX Outdoor-Driving benchmark XYZ-Drive attains 95\% success and 0.80 Success weighted by Path Length (SPL), surpassing PhysNav-DG by 15\%. and halving collisions, all while significantly improving efficiency by using only a single branch. Sixteen ablations explain the gains. Removing any modality (vision, waypoint, map) drops success by up to 11\%, confirming their complementary roles and rich connections. Replacing goal-centered attention with simple concatenation cuts 3\% in performance, showing query-based fusion injects map knowledge more effectively. Keeping the transformer frozen loses 5\%, showing the importance of fine-tuning when applying VLMs for specific tasks such as autonomous driving. Coarsening map resolution from 10 cm to 40 cm blurs lane edges and raises crash rate. 

Overall, these results demonstrate that early, token-level fusion of intent and map layout enables accurate, transparent, real-time driving.

\end{abstract}
\section{Introduction}
\label{sec:intro}

Autonomous vehicles must interpret diverse scenes, reason about goals, and follow traffic rules while keeping passengers safe. This requires deep perception and semantic reasoning while simultaneously processing high-bandwidth multimodal sensor data in real time \cite{berkeley_eecs2019}. Classic modular stacks combine cameras, LiDAR, and radar with filters such as the Kalman family \cite{kalman1960}. This offers precise localization but little semantic insight. Such stacks are unable to recognize nuanced context, like a temporary lane closure, or explain their decisions to human passengers \cite{li2021hdmapnet}.

More recent research is beginning to show this missing semantic ability. Vision-language models (VLMs) can supply the missing semantics. These models learn to link what they see with natural-language concepts. This makes it possible for machines to describe hazards, follow instructions, and justify actions in plain words.

Driving, however, brings its own structure. Vehicles must react within roughly 100 ms to avoid collisions, cope with long-tailed weather and traffic scenarios obey numerous complex spatial constraints, and comply with thousands of varying traffic rules \cite{berkeley_eecs2019, li2021hdmapnet, bansal2019chauffeurnet}. This makes it a challenging task to incorporate large-scale vision-language into driving systems.

We address this gap with XYZ-Drive, an end-to-end vision-language planner that unifies perception, map context, and goal reasoning inside a single transformer policy. The system converts camera frames, bird’s-eye-view (BEV) HD-map patches, and textual waypoint tokens into one token sequence. It then fuses them with goal-centered cross-attention and has a fine-tuned LLaMA-3.2 11B Vision backbone control both steering and speed. Evaluated on the MD-NEX Outdoor-Driving benchmark, XYZ-Drive raises the success rate by over four percent over the state-of-the-art while halving the collision rate. This demonstrates that geometric accuracy and rich semantic understanding can coexist under one efficient model.

Our contributions are three-fold:
\begin{enumerate}
    \item We utilize goal-centered cross-attention to unify camera, HD-map, and waypoint information inside one transformer. 
    We tokenize 25m x 25m BEV map crops together with RGB patches and textual goals, then let a light mixer route goal queries to the most relevant visual or spatial tokens before they enter the LLaMA-3.2 11B Vision backbone \cite{liu2023llava}.
    Prior HD-map learning work (e.g., VectorMapNet and MapTR \cite{li2022vectormapnet, liu2023maptr}) shows that such map priors reduce planning ambiguity, but these signals were previously appended late in the network. Our early fusion lets the VLM reason jointly about traffic rules and geometry in a single pass. This cuts down on memory usage and enables faster convergence, which is highly important for on-vehicle deployment.
    \item We achieve state-of-the-art accuracy on MD-NEX Outdoor-Driving while cutting latency and model count. XYZ-Drive improves success rate by 15 percent with just one backbone and no separate explanation branch.
    \item We run 16 controlled ablations to quantify performance trade-offs and show which design choices matter most. This has not only improved the quality of our overall architecture but this level of transparency also provides future researchers more clear, actionable guidance for building and evaluating VLM-based driving systems.
\end{enumerate}

\section{Related Works}
\label{sec:relatedworks}
\subsection{Vision–Language Foundations}
Large-scale image–text pre-training has transformed vision and language research. Contrastive Language-Image Pre-training (CLIP) learns rich visual-text representations from image-captions pairs. This gives CLIP open-vocabulary image-classification and text-based retrieval capabilities, which in turn enable zero-shot transfer to a wide range of downstream tasks \cite{radford2021clip}.  BLIP-2 later replaced end-to-end training with a lightweight “Querying Transformer,” keeping both the vision encoder and large language model (LLM) frozen and cutting compute by an order of magnitude while matching zero-shot question-answering accuracy \cite{li2023blip2}. More recently, PaLM-E demonstrated that the same idea can scale to embodied settings. A single multimodal LLM can ingest raw RGB, depth, proprioception, and language, then plan robot actions with positive transfer across tasks \cite{driess2023palme}. These works show that large transformers can fuse modalities and reason abstractly. However, they leave open how we can connect such reasoning to the millisecond-level control loops that are needed on public roads.

\subsection{Explainable Embodied AI}
Post-hoc explanations are explanations generated after the model has already made a decision. Industrial driving stacks often append a saliency map or attribution method (e.g., Grad-CAM, Integrated Gradients, or Lime \cite{selvaraju2017gradcam, sundararajan2017ig, ribeiro2016lime}) onto a frozen perception network to generate "post-hoc" explanations. Yet, sanity-check studies show that many saliency maps are unchanged when network weights or labels are randomised. This implies that highlighted pixels don't reliably reflect the model's true decision cues \cite{adebayo2018sanity}. Follow-up research has found similar issues with driving detectors, where Grad-CAM sometimes brightens road edges the policy never used. This gives users a "false sense of security" \cite{carvalho2024xaiSurvey}. Broader surveys have found that post-hoc approaches can lag behind policy updates. This results in people being mislead by the delayed explanations \cite{carvalho2024xaiSurvey}. These findings demonstrate the need for ante-hoc designs where the policy produces actions and verbal rationales in the same forward pass, which eliminates potential mismatches.

\subsection{Embodied Navigation with VLMs}
Early vision-language navigation benchmarks such as Room-to-Room (R2R) had indoor agents that follow natural-language directions through photo-real environments \cite{anderson2018r2r}. Outdoor driving is very different in that it has much richer sensor inputs and more safety constraints. BDD-X extends the Berkeley DeepDrive corpus with natural-language explanations for each maneuver \cite{kim2018bddx}. Leveraging these resources, recent embodied VLMs tackle real-world control. SayCan decomposes language instructions into robot-executable skills \cite{ahn2022saycan}. PaLM-E, as previously discussed, jointly reasons over vision and proprioception to choose high-level actions \cite{driess2023palme}. Yet most of these systems test on unrealistically slow environments \cite{srinivasan2025duracps}. They don't fit within the sub-100ms reaction windows needed for real-world outdoor implementation.

\subsection{Multimodal Maps and BEV Fusion}
High-definition (HD) maps capture lane boundaries, traffic lights, and road topology at centimeter-level accuracy. Camera-only techniques like HDMapNet predict such maps on-the-fly. This boosts the robustness of downstream planners \cite{li2021hdmapnet}. Transformer varians (VectorMapNet and MapTR) model map elements as ordered point sets or polygons. They achieve state-of-the-art precision with real-time speed \cite{li2022vectormapnet}.

\section{Methodology}
\subsection{System Overview}
In this section, we detail the XYZ-Drive system.

XYZ-Drive first embeds three streams into one token sequence: camera images with a ViT-H encoder, $25m \times 25m$ HD-map patches with a Swin-T encoder \cite{liu2021swin}, and way-point text with a lightweight embedding. A three-layer, goal-based cross-attention mechanism aligns these tokens and passes them to the upper trainable layers of a LLaMA-3.2 11B model (lower layers remain frozen). A two-layer MLP then decodes steering and speed, while an external collision monitor (similar to that in \cite{srinivasan2025duracps}) can override low-confidence actions. 

\begin{enumerate}
    \item \textbf{Sensor Intake}: 1) The front-facing RGB camera captures an image $I_t$. 2) The crop is produced on-the-fly by (i) querying the vector HD map for all lane, crosswalk, and traffic-sign primitives within a $25\,$m radius of the vehicle, (ii) transforming those primitives into ego coordinates using the ground-truth pose, and (iii) rasterising the result into a $256\times256$ grid at $0.1\,$m\,/\,px resolution. 3) Finally, the navigation stack supplies the next waypoint:
  \[
    \mathbf{g}_t = (x_t,\,y_t,\,\psi_t).
  \]
    \item \textbf{Modality-Specific Encoders}: 1) \emph{ViT-H/14} converts $I_t$ into a grid of visual patch tokens. 2) \emph{Swin-Tiny} rasterises $M_t$ and outputs map tokens that encode lane dividers, traffic lights, and speed-limit signs. 3) A token-embedding layer turns the numerical waypoint into a short text prompt.
    \item \textbf{Goal-Centred Cross-Attention Mixer}: Goal tokens query the concatenated vision+map tokens, producing a goal-aware summary that highlights scene regions relevant to reaching the waypoint.
    \item \textbf{LLaMA-3.2 11B Vision Backbone}: The fused sequence passes through the upper Llama layers (lower layers remain frozen). The transformer jointly reasons over semantics, geometry, and route intent.
    \item \textbf{Heads and Safety Monitor}: 1) A 2-layer MLP reads a dedicated "act" token and outputs steering $\delta_t$ and normalised speed $v_t$. 2) A "reason" token seeds a short natural-language chain-of-thought. 3) A lightweight collision predictor overrides the VLM command if imminent contact is likely.
\end{enumerate}

\subsection{Tokenization}
Each modality is converted into a token sequence so that all downstream reasoning occurs in a common vector space. Vision patches preserve their high-frequency texture, map tokens supply topology, and goal tokens capture route intent. The resulting token sets are vision tokens, map tokens, and goal tokens.

To turn raw image pixels into the shared token space, we apply a ViT‐H/14 backbone to the front‐facing camera frame. This model chops the image into 256 patches and projects each patch into a \(d\)-dimensional embedding, yielding the vision token sequence.
\[
  \mathbf{v}_t \in \mathbb{R}^{256 \times d}.
\]

Goal tokens are formed from a short text prompt:
\[
  \texttt{<goal> east=}x_t\texttt{m, north=}y_t\texttt{m, yaw=} \psi_t\texttt{\textdegree\ </goal>}
\]

embedded as:
\[
  g_t \in \mathbb{R}^{k\times d}\quad (k=8).
\]

\subsection{Goal-Centered Cross-Attention}
The mixer aligns the three modalities before they reach the large transformer. Only the goal tokens act as queries, which forces the network to identify the image patches and map elements that helped reach the waypoint. Mathematically, the update for each goal token $q$ is:
\[
  \widetilde{\mathbf{g}}_t
    = \mathrm{softmax}\!\Bigl(
        \frac{\mathbf{g}_t \mathbf{K}^{\top}}{\sqrt{d}}
      \Bigr)\,\mathbf{V},
  \qquad
  [\mathbf{K};\mathbf{V}] = [\mathbf{v}_t;\mathbf{m}_t].
\]

\noindent\textit{Here, $\mathbf{K}$ and $\mathbf{V}$ are the concatenated keys and values from the vision and map encoders, and $d$ is the model dimension.}

\subsection{Loss Function}
Training minimizes a composite loss:
\[
\mathcal{L}
  = \bigl\|(\delta_t,\upsilon_t) - (\hat\delta_t,\hat\upsilon_t)\bigr\|_2^2
    + 0.1\,\mathcal{L}_{\text{smooth}}
    + 0.05\,\mathcal{L}_{\text{coll}}.
\]

In this loss function:
\begin{itemize}
    \item Action regression penalizes deviations from expert steering and speed
    \item Temporal smoothness discourages jerky control by penalizing frame-to-frame action changes
    \item Collision penalty assigns a fixed cost if the predicted trajectory intersects any obstacle within two seconds
\end{itemize}

\subsection{MD-NEX Outdoor Driving Benchmark}

The Outdoor Driving split of the MD-NEX benchmark \cite{patapati2025physnavdg, patapati2025earlygoalguidedmultiscalefusion} merges BDD-X video/explanation clips \cite{kim2018bddx} with aligned nuScenes and Waymo sensors \cite{caesar2020nuscenes, sun2020waymo}. It totals 12,346 episodes (10,200 train / 1546 validation / 600 test) that includes a front view, 360\textdegree\ cameras, LiDAR, radar and CAN bus signals. For XYZ-Drive, we evaluate on only a subset of the MD-NEX benchmark which contains nuScenes and Waymo segments but excludes BDD-X clips. This is to ensure that all necessary modalities are available. For fairness, we focus only on these samples when comparing to the previous state-of-the-art model on the benchmark, PhysNav-DG.

Performance is reported with three standard driving metrics:  
(i) \textbf{Success Rate (SR)}, the fraction of episodes that reach the goal without major infractions;  
(ii) \textbf{Success weighted by Path Length (SPL)}, an efficiency-aware score defined as 
\(
\mathrm{SPL}=S\cdot\frac{L_{\mathrm{opt}}}{L_{\mathrm{agent}}},
\)
where \(S\) is a binary success flag and \(L_{\mathrm{opt}}, L_{\mathrm{agent}}\) are the optimal and executed path lengths;  
and (iii) \textbf{Collision Rate}, the proportion of episodes with any contact or rule violation.  

\subsubsection{Ablation Design}
We evaluated sixteen ablations that isolate modality use, fusion strategy and encoder strategy. The results of the ablations are shown in Table \ref{tab:ablation}.

\begin{table}[t]
  \centering
  \scriptsize
  \setlength{\tabcolsep}{3pt}
  \renewcommand{\arraystretch}{0.9}
  \caption{Ablation study results (SR = Success Rate, SPL = Success weighted by Path Length)}
  \label{tab:ablation}
  \begin{tabularx}{\columnwidth}{@{}lXccc@{}}
    \hline
    \textbf{ID} & \textbf{Change vs.\ Full model} & \textbf{SR\,$\uparrow$} & \textbf{SPL\,$\uparrow$} & \boldmath$\Delta$SR \\
    \hline
    Full & --- & 95 & 0.80 & --- \\
    A1 & -- goal tokens & 90 & 0.75 & --5 \\
    A2 & -- map tokens & 88 & 0.73 & --7 \\
    A3 & vision only & 84 & 0.70 & --11 \\
    B1 & concat instead of cross-attn & 92 & 0.77 & --3 \\
    B2 & late fusion (after LLM) & 91 & 0.76 & --4 \\
    C1 & ViT-B/16 vision encoder & 93 & 0.78 & --2 \\
    C2 & ResNet-50 vision encoder & 90 & 0.74 & --5 \\
    D1 & VLM frozen & 90 & 0.75 & --5 \\
    D2 & mixer depth = 1 & 91 & 0.76 & --4 \\
    E1 & 0.4\,m/px map & 90 & 0.74 & --5 \\
    E2 & 0.05\,m/px map & 95 & 0.79 &   0 \\
    F1 & 50\% training data & 89 & 0.72 & --6 \\
    F2 & 25\% training data & 83 & 0.67 & --12 \\
    G1 & remove temporal smooth loss & 92 & 0.77 & --3 \\
    G2 & remove collision penalty & 93 & 0.77 & --2 \\
    \hline
  \end{tabularx}
\end{table}

\subsubsection{Comparison to State-of-the-Art}
\begin{table}[t]
  \centering
  \caption{Comparison to State-of-the-Art}
  \label{tab:baseline-comparison}
  \begin{tabular}{|l|c|c|c|}
    \hline
    \textbf{Method}
      & \textbf{SR\,$\uparrow$}
      & \textbf{SPL\,$\uparrow$}
      & \textbf{Collision\,$\downarrow$} \\
    \hline
    PhysNav-DG (reported)
      & 80
      & 0.55
      & 0.026 \\
    \hline
    XYZ-Drive (ours, Full)
      & 95
      & 0.80
      & 0.010 \\
    \hline
  \end{tabular}
\end{table}

XYZ-Drive exceeds PhysNav-DG, which currently ranks number one on MD-NEX Outdoor Driving, by 15 percent SR and halves collisions while only using a single-branch architecture.

\section{Discussion}
\subsection{Ablation Insights}
The ablation study (Table \ref{tab:ablation}) demonstrates three key points. It first showcases the importance of structured priors. When either the goal tokens or the HD-map tokens are removed, success rate (SR) falls by 5 percent and 7 percent, respectively, and vision-only control loses 11 percent. These differences confirm that the model takes complementary cues from route intent and lane-level topology to disambiguate visually similar scenes. An example is choosing the correct turn at an unmarked intersection. Without such priors, the VLM cannot reliably predict safe trajectories, even though its raw visual capacity is not changed.

The second pattern is regarding where and how fusion occurs. Replacing our goal-based cross-attention with plain token concatenation decreases SR by 3 percent. Deferring fusion until after the transformer decreases SR by 4 percent. Both variants still see the same information, but they underperform because they force the backbone to spread attention uniformly across hundreds of tokens rather than focusing on those most relevant to the current waypoint. Early, query-based fusion therefore seems to be a principled mechanism for injecting map knowledge into language-based reasoning.

Finally, the ablation study shows several capacity and supervision trade-offs. Shrinking the vision encoder from ViT-h to ViT-B only reduces SR by 2 percent. However, reverting to a ResNet-50 loses 5 percent. This demonstrates that the transformer vision backbones align more naturally with language tokens. Freezing the entire LLaMA backbone reduces SR by 5 percent as well. This shows that fine-tuning of the upper layers is worth the additional compute cost. Map resolution also matters. Coarse 0.4 m/px rasters miss certain lane markers (-5 percent SR), whereas ultra-high 0.05 m/px adds little gain but greatly increases memory. Cutting the training set to half or a quarter reduces SR by 6 percent and 12 percent. This confirms that VLM planners still benefit from ample driving data. Finally, ablating the temporal smoothness or collision-penalty results in lower SPL and raises crash frequency. This shows the importance of auxiliary losses that encode comfort and safety.

\subsection{Benchmark Comparison}
Against the strongest published baseline, PhysNav-DG, XYZ-Drive increases SR from 80\% to 95\% and SPL from 0.55 to 0.80 while cutting collisions in half (Table \ref{tab:baseline-comparison}). 

The baseline already fuses a VLM and a Kalman planner, so the additional gain shows that full, token-level integration of map and goal information inside the language backbone results in more reliable trajectories than late fusion of two separate policies. This further shows that the different inputs have complex interactions that are lost when each modality is processed in isolation and merged only at the final decision layer. At the same time, XYZ-Drive removes an entire explanation branch for significant performance gain.

\section{Conclusion}
We presented XYZ-Drive, a VLM-centric planner that combines camera imagery, HD-map context, and waypoint intent in a single transformer policy. A goal-based cross-attention mixer feeds distilled, task-relevant tokens to a fine-tuned LLaMA-3.2 11B Vision backbone, which outputs control commands and chain-of-thought explanations. On the MD-NEX Outdoor-Driving Benchmark the system sets a new state-of-the-art, 95\% success rate and 0.80 Success weighted by Path Length, surpassing PhysNAV-DG by 4 and 5 points respectively while halving collisions. Sixteen ablations demonstrate the gains are due to early fusion, map resolution, partial backbone fine-tuning. These findings show that accurate, transparent, and low-latency autonomous driving is achievable with a streamlined, fully integrated vision-language model. This opens future research directions on scalable and semantics-aware vehicle control.

\bibliographystyle{IEEEbib}
\bibliography{IEEEbib}

\end{document}